# Mining Local Gazetteers of Literary Chinese with CRF and Pattern based Methods for Biographical Information in Chinese History

**Chao-Lin Liu**[†]     **Chih-Kai Huang**[§]     **Hongsu Wang**[‡]     **Peter K. Bol**[!]

[†§]Department of Computer Science, National Chengchi University, Taiwan
[‡!]Institute for Quantitative Social Science, Harvard University, USA
[†]Graduate Institute of Linguistics, National Chengchi University, Taiwan
{[†]chaolin,[§]102753029}@nccu.edu.tw, {[‡]hongsuwang,[!]pkbol}@fas.harvard.edu

*Abstract*—Person names and location names are essential building blocks for identifying events and social networks in historical documents that were written in literary Chinese. We take the lead to explore the research on algorithmically recognizing named entities in literary Chinese for historical studies with language-model based and conditional-random-field based methods, and extend our work to mining the document structures in historical documents. Practical evaluations were conducted with texts that were extracted from more than 220 volumes of local gazetteers (*Difangzhi*, 地方志). *Difangzhi* is a huge and the single most important collection that contains information about officers who served in local government in Chinese history. Our methods performed very well on these realistic tests. Thousands of names and addresses were identified from the texts. A good portion of the extracted names match the biographical information currently recorded in the China Biographical Database (CBDB) of Harvard University, and many others can be verified by historians and will become as new additions to CBDB.[1]

## I. INTRODUCTION

Person and location names are crucial ingredients for studying historical documents. Knowing the participants and locations provides a solid foundation for analyzing historical events. Detecting temporal markers is also very important for historical studies, yet, for Chinese history, it is relatively easier to spot the temporal markers because the names of the dynasties and reign periods (年號, nian2 hao4) are known and stable.

We apply techniques of textual analysis and machine learning to find person names, location names, and their relationships in *Difangzhi* (地方志, **DFZ** henceforth) in the present work, aiming to enrich the contents of the China Biographical Database [1]. DFZ is a general name for a large number of local gazetteers that were compiled by local governments of different levels in China since as early as the 6th century AD (cf. [7]). A 1995 study reported that there are at least more than 8200 series of DFZ found in China ([13], p. 11). DFZ contain a wide range of information about the locality, and the biographical information about the government officers is our current focus.

The main barrier for achieving our goals is that there is little completed work in the literature about the grammars of literary Chinese, although grammars are central for extracting named entities like person and location names from texts with computational methods [6][18].

Figure 1 shows the image of a sample DFZ page. In the old days, Chinese texts were written from top to bottom and from right to left on a page. Most linguists know that there are no word boundaries in modern Chinese. It might be quite surprising for researchers outside of the Chinese community that there were even no punctuations in literary Chinese. Without clear delimiters between words and sentences, it is very challenging even for people to read literary Chinese, so it takes research to find ways to segment words and split sentences in literary Chinese [9].

Grammar induction [5] is a general name for enabling computers to learn the grammars of natural languages. Some researchers have worked on the grammars for selected sources of Chinese. Huang et al. [10] explored the induction problem with about a thousand sentences that were extracted from Hanfeizi (韓非子) and Xunzi (荀子), philosophical texts from over two thousand years ago. Kuo [12] has tried to

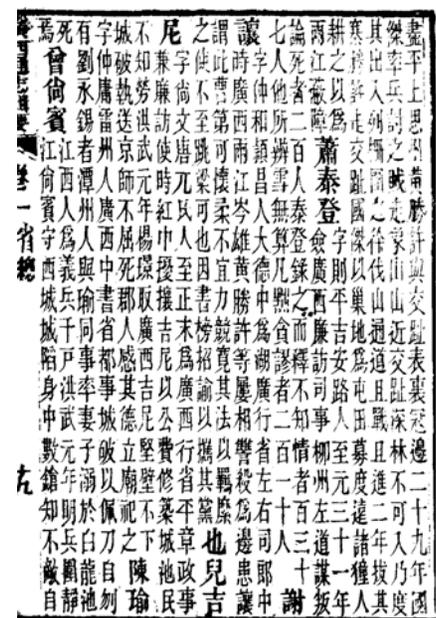

**Figure 1. A page of DFZ**

---

find phrase-structure rules for modern Chinese texts, and Lee and Kong [14] have built treebanks for Tang poems. Although these researchers have worked on grammars for Chinese texts, they encountered Chinese statements that are quite different from the ones that we need to handle in DFZ.

Previous work for inducing grammars of literary Chinese employed some forms of pre-existing information to begin the induction procedures. Given that literary Chinese texts consist of long sequences of characters, the needs for external information for grammar induction should be expected. Hwa [11] assumed that the training corpus was partially annotated with high-level syntactic labels. Lü et al. [16] started with bilingual corpora. Yu et al. [22] undertook their work with a sample treebank, and Boonkwan and Steedman [4] began with some syntactic prototypes.

We tackle the NER tasks in literary Chinese from two unexplored perspectives. First, we employ the biographical information in the China Biographical Database (**CBDB**, henceforth) to annotate the DFZ texts, learn language models (**LMs**, henceforth) from the annotated texts, and extract biographical information based on the learned models. Alternatively, we train conditional-random-field [20] models with a set of labeled DFZ data that were achieved by Bol and his colleagues [1][19], and use the conditional-random-field (**CRF**, henceforth) models to extract candidate names from the test data, which is another set of DFZ texts. We have verified the findings of the LM-based and the CRF-based methods. Both show very good results for NER in DFZ.

Furthermore, we used the identified names to separate continuous strings of Chinese characters into paragraphs so that we can more precisely know the career of the subject persons as recorded in the paragraphs.

We present the sources of our data, define our target problems, and discuss the motivation for our work in Section II. We then provide details about our main approaches in two long sections. In Sections III and IV, we look into details about the LM-based and CRF-based methods, respectively, including the designs of the classification models and results of several evaluation tasks. In Section V, we offer our current results of separating the continuous text statements in DFZ into paragraphs. In Section VI, we conclude with a brief summary and discussions of some technical issues.

## II. DATA SOURCES, PROBLEM DEFINITIONS, AND MOTIVATION

We provide information about the sources of our data, define the problems that we wish to solve, and explain the rationale of our approach in this section.

### A. Unlabeled Data

Currently, we have two sets of DFZ text files. The unlabeled part has more than 900 thousand characters that were extracted from 83 volumes of local gazetteers [3]. The labeled part will be presented in Section II.D.

These 83 volumes were compiled between the middle of the Ming dynasty (1368-1644AD) and the early Republican period (1912-1949). These books were produced by governments of different levels at 65 locations in China.

Figure 1 shows a sample page from this collection. It is hard to count the number of columns on this page. Typically, we consider one line of characters as a column but here we see that a column is split into two columns of smaller characters. The legal name is emphasized by having it occupy the full width of a column, and details about this person are recorded in the thinner sub-columns. When the leftmost three columns of Figure 1 are written out as a continuous horizontal text they read as shown in Figure 2 in which the full-width person names are in bold type.

The DFZ texts may contain characters that are not or seldom used in modern Chinese and are missing from the Unicode. If these characters have modern counterparts, they will be substituted for by their modern replacements; otherwise, spaces will take their positions. As an example of the former case, the thirteenth character on the first column from the right in Figure 1 is "裏" (li3), which may be written as "裡" (li3) in our files. When the latter case occurs understanding the original DFZ records becomes even more challenging.

### B. Problem Definitions

We wish to build a system that can extract biographical information from DFZ to enrich the contents of CBDB. The current contents of CBDB were extracted from sources other than DFZ [19]. Hence, we are interested in spotting all types biographical information in DFZ.

In this paper, we focus on issues about finding person names and location names, and extend to some relevant topics, such as checking whether the locations were birth places and finding paragraph boundaries. In the long run, we will expand to social networks and personal careers as well.

*C. More on Motivation*

In a text passage such as Figure 2, it is very challenging to find useful information without assistive information, even for modern generations of native speakers of Chinese. Without the bold type it would not be easy to find the name "陳瑜" (chen2 yu2) which was written in larger characters in the original DFZ.

The grammars of literary and modern Chinese are not exactly the same, and reading literary Chinese is a lot harder than reading modern Chinese, especially when there are no boundary markers between sentences. In addition, historical knowledge is also required for correct word segmentation and lexical disambiguation, which are important for understanding and extracting the desired information from the texts.

不知勞洪武元年楊璟取廣西吉尼堅壁不下城破執送京師不屈死郡人感其德立廟祀之**陳瑜**字仲庸雷州人廣西中書省都事城破以佩刀自刎有劉永錫者潭州人與瑜同事率妻子溺於白龍池死焉**曾尚賓**江西人為義兵千戶洪武元年明兵圍靜江尚賓守西城城陷身中數鎗知不敵

**Figure 2. A partial DFZ passage from Figure 1**

To achieve our goals, we need some informative sources. The importance of these informative sources for our methods for extracting information is just like that of the machine-readable dictionaries for the methods for handling modern natural languages.

Our approaches are innovative because we utilize the biographical information in CBDB to provide semantic information about the DFZ texts. In contrast, the literature that we reviewed in Section I carried out grammar induction with such linguistic knowledge as part-of-speech tags and syntactic structures.

*D. Labeled Data*

We have a set of labeled DFZ data. This set of data was collected from 143 volumes of DFZ, which contain more than 1.498 million characters.

The DFZ texts were labeled with regular expressions (**REs**, henceforth) that were compiled by domain experts [1][19], and the REs were designed to extract biographical information. The labeled data were then saved as records in a large table with 113,784 records in total.

Each record has many fields, and the fields were designed to contain a wide variety of factoids about the individuals. Major fields contain information about an individual's official name, style name (字, zi4), pen name (號, hao4), dynasty, native place (籍貫, ji2 guan4), office (官職, guan1 zhi2), entry method (入仕方法, ru4 shi4 fang1 fa3), dates of tenure in office, office location, and reign period (年號, nian2 hao4).

Due to the nature of the original DFZ data and the limited expressiveness of REs, a non-negligible portion of the fields do not have values (i.e., have missing values), and, sometimes, the values are not correct. Nevertheless, these labeled data remain valuable and prove to be useful from the perspective of historical studies [19] and of building machine-learning models.

### III. LANGUAGE-MODEL-BASED APPROACH

We annotate DFZ with the biographical information available in CBDB, and find the frequent and *consistent* n-grams for locating candidate strings from which we may extract official names and style names.

*A. Labeling and Disambiguation*

Figure 3 lists the steps of our main procedure, **Constrained N-Grams** (**CNGRAM**), for NER. First, we label the text with biographical information in CBDB. Five types of labels are in use now: **name** for an official or a style name, **address** for locations, **entry** for entry methods, **office** for service office, and **nianhao** for reign periods.

In reality, some strings may be labeled in more than one way. For instance, "陽朔" (yang1 shuo4) can refer to a reign period of the Han dynasty or a location name, and "王臣" (wang2 chen2) is a very popular official name that was used in many dynasties. Before we try to disambiguate the labeling, we keep all possible labels for a string.

We will use the following short excerpt from Figure 2 to explain the execution of CNGRAM.

> **T1:** 陳瑜字仲庸雷州人廣西中書省都事

Identifying T1 from its context is possible because this string begins and ends with words that have corresponding labels. "陳瑜" (chen2 yu2) was the name of several officials from across the Yuan, Ming, and Qing dynasties, and both "雷州" (lei2 zhou1) and "廣西" (guang3 xi1) were addresses.

In the first step of CNGRAM, we prefer longer matches for the same type of labels as a heuristic principle for disambiguation. The principle of preferring longer words is very common for Chinese word segmentation. In T1, both "中書省都事" (zhong1

shu1 sheng3 dou1 shi4) and "中書省" can be labeled as office names in the Yuan dynasty, but we would choose only the former because "中書省都事" is a longer string. In contrast, we do not have "中書省都事" for the Ming dynasty, so will use "中書" and "都事" for Ming.

We also assume that named entities in a passage should be *consistent* in some sense as another heuristic principle for disambiguation. This consistent principle is reminiscent of the "**one sense per discourse**" principle for word sense disambiguation [21] in natural language processing.

Currently, we presume that named entities in a context of six labels should be referring to something of the same dynasty, where six is an arbitrary choice and can be varied. We have not used addresses to check consistency because we are still expanding our list of addresses. Therefore, we do not accept a "陳瑜" of the Qing dynasty because neither "中書省都事" nor "中書" was an entity in Qing. Using the consistent principle, we will keep labels only for the Song and Ming dynasties for the sample passage, thereby achieving some disambiguation effects.

**Procedure CNGRAM** (txt, idbs, cc)
   *txt*: DFZ texts
   *idbs*: information databases
   *cc*: chosen conditions for checking consistency
**Steps**
1. Label the text based on the given *idbs*. Prefer the labels that cover longer strings, all else being equal.
2. For contexts of chosen conditions, *cc*, remove the inconsistent labels.
3. Find the frequent consistent *n*-gram patterns, and use them as filter patterns
4. Try to extract named entities from strings that conform to the filter patterns

**Figure 3. The CNGRAM procedure**

Hence, we have two consistent sequences for T1: [name("陳瑜", Yuan), address("雷州"), address ("廣西"), office("中書省都事", Yuan)] and [name("陳瑜", Ming), address("雷州"), address("廣西"), office("中書", Ming), office("都事", Ming)].

## B. Extracting Unknown Style Names

Aiming at extracting person and style names for government officers, we focus on the consistent sequences that have at least one **name** label. After labeling texts with CNGRAM, we identify and prefer strings that are associated with more different labels. We show four such *filter patterns* below.

   **P1: name-address-nianhao-entry**
   **P2: name-address-entry-nianhao**
   **P3: name-name-address-address**
   **P4: name-address-address-office**

These patterns shed light on how person names were presented in DFZ texts. We can now examine the DFZ strings that are labeled with these patterns to judge whether these patterns indeed carry useful information. Usually, we find regularities in these statements, and can implement specific programs for extracting target information from such patterns.

Our running example, T1, follows the P4 pattern in two different ways, and we list the substrings.

   **T2:** 陳瑜字仲庸雷州人廣西中書
   **T3:** 陳瑜字仲庸雷州人廣西中書省都事

In both T2 and T3, we see that a key signal "字" (zi4), which is a typical prefix for style names, follows a **name** label. "字" is followed by two unlabeled characters which are then followed by an **address**, an unlabeled character, another **address**, and an **office**. Thus, T2 and T3 are examples of pattern P5, where <name> and <address> represent labeled strings and Z1 and Z2 are two unlabeled characters.

   **P5: <name>　字　Z1　Z2　<address>**

The unlabeled characters, Z1 and Z2, can be extracted as style names because practical statistics indicate that over 98% of style names contain exactly two characters. Therefore, we embody this finding with actions in our programs.

The third step in CNGRAM is thus an interactive step[2], and requires human participation. Notice that the work for domain experts is minimal and that the results are worthwhile. A human expert does not have to read 83 books to find the candidate patterns. Using CNGRAM to locate string patterns that contain useful information, we are able to process a large amount of data both efficiently and effectively.

With the extracted style name "仲庸" (zhong4 yong1), we can create two records, i.e., (Yuan, 陳瑜, 仲庸) and (Ming, 陳瑜, 仲庸). "仲庸" is unknown to CBDB, and can be added to CBDB with the approval of domain experts.

---

[2] Using computers to select the patterns is possible if we are willingly to set a frequency threshold to determine "frequent" patterns, which may not be a perfect choice for historical studies.

The CNGRAM procedure actually helps us learn the language models that were used in DFZ. By inspecting frequent and consistent patterns that actually contain biographical information, we can gather more knowledge about grammar rules in DFZ and then implement NER procedures based on the observations.

*C. Empirical Evaluations*

We compared the extracted records with the records in CBDB (2014 version) to evaluate the CNGRAM procedure, and show the results in Table 1, where the circles and crosses, respectively, indicate matches and mismatches between the extracted and CBDB records.

The matching results are categorized into seven types, e.g., type 1 is the group that we had perfect matches in dynasty, official name, and style name. We have 609 such instances in the current experiment, and the proportion of type-1 instances is 28.3% of the 2152 extracted records.

The two records that we obtained in the previous subsection belong to type 2, because "仲庸" is not known to CBDB. All extracted records of type 2 provide opportunities of finding unknown style names for CBDB. However, they should be confirmed by historians. The experts may check the original texts for this approval procedure, which is an operation facilitated by our software platform.

Table 1. Analysis of 2152 extracted records

| Type | Dynasty | Name | Style N. | Quan. | Prop. |
|---|---|---|---|---|---|
| 1 | ○ | ○ | ○ | 609 | 28.30% |
| 2 | ○ | ○ | × | 665 | 30.90% |
| 3 | × | ○ | ○ | 117 | 5.44% |
| 4 | ○ | × | ○ | 262 | 12.17% |
| 5 | × | ○ | × | 220 | 10.22% |
| 6 | × | × | ○ | 45 | 2.09% |
| 7 | ○/× | × | × | 234 | 10.87% |

○不知勞洪武元年楊璟取廣西吉尼堅壁不下○城破執送京師不屈死郡人感其德立廟祀之○陳瑜○字仲庸雷州人廣西中書省都事城破以佩刀自刎○有劉永錫者潭州人與瑜同事率妻子溺於白龍池○死○焉○曾尚賓○江西人為義兵千戶洪武元年明兵圍靜○江尚賓守西城城陷身中數鎗知不敵自○

**Figure 4. A partial DFZ passage with circles**

Records of types 3 and 4 are similar to records of type 2. They offer opportunities to add extra information to CBDB. Records of types 5, 6, and 7 provide some opportunities for adding information about new persons to CBDB. After inspecting the original text segments, we will be able to tell whether these mismatches are new discoveries or just incorrect extractions.

*D. Further Extensions*

We are more ambitious than verifying whether CNGRAM can help us find correct biographical information. Type-1 records can be instrumental for advanced applications. They help us find the beginnings of the descriptions that contain information about the owners of type-1 records.

If we can determine the beginnings of two consecutive segments, then we can find persons who have relationships. T1 is the beginning of a major segment in Figure 1. The string "也兒吉尼字尚文唐兀氏人" is the beginning of a segment for a person named "也兒吉尼" (ye3 er2 ji2 ni2). The person mentioned between "也兒吉尼字尚文唐兀氏" and T1, e.g., "楊璟" (yang2 jing3), should have some relationships with "也兒吉尼". We will present results of this line of work in Section V.

In addition, it is quite intriguing to apply pattern P5 (Section III.B) in an extreme way. Figure 4 shows the raw data for the text in Figure 2. If we compare Figure 4 and the image in Figure 1 carefully, we can find that the circles were added to signify (1) changes between major columns and thin sub-columns or (2) line breaks. The semantics of the circles are ambiguous, but they are still potentially useful.

If "字" is really a strong indicator that connects legal names and their style names, P6 and P7 may lead us to find pairs of official and style names. Here, we use C1, C2, and C3 to denote Chinese characters.

    **P6:** ○   C1   C2   C3 字   Z1   Z2
    **P7:** ○   C1   C2   字   Z1   Z2

When we find substrings that conform to P6 or P7 in the raw data, we may want to check whether C1-C2-C3 (or C1-C2) is an official name, Z1-Z2 is a style name, and their combination is for a real person.

We evaluated this heuristic approach with the unlabeled data of Section II.A, and found 3765 pairs of (legal_name, style_name). We checked these pairs with CBDB (2014 version), and achieved Table 2. Because strings conforming to P6 and P7 have very short contexts, we could not judge the dynasties of these names, so Table 2 is simpler than Table 1.

Table 2 shows that 31% of the pairs have corresponding records in CBDB. Although we must inspect the original texts to verify the correctness of these matched records, the statistics are promising and encouraging. 1192 type-1 records matched the official and style names of certain CBDB records. This amount is more than the number of type-1 records in Table 1. Some of the pairs that we identify with the current heuristic did not appear in filter patterns that we discussed in Section III.B, suggesting that a hybrid approach might be worthy of trying in the future.

**Table 2. Analysis of 3765 extracted records**

| Type | Name | Style Name | Quan. | Prop. |
|---|---|---|---|---|
| 1 | ○ | ○ | 1192 | 31.66% |
| 2 | ○ | × | 885 | 23.51% |
| 3 | × | ○ | 1104 | 29.32% |
| 4 | × | × | 584 | 15.51% |

**Table 3. Features for CRF models**

| Group | Types | Description |
|---|---|---|
| 1 | Chinese characters | self |
| 2 | Chinese characters | surrounding $k$ characters |
| 3 | relative positions of selected named entities | office, entry, reign period, and time |
| 4 | usage | used in person or location name |
| 5 | usage | family name? |
| 6 | named entities | office, entry, reign period, and time |

## IV. CRF-BASED APPROACH

CRF-based models [20] are very common for handling NER with machine learning methods [18]. We employed MALLET [17] tools for building linear-chain CRF models, and trained and tested our models with the data that we discussed in Section II.

### A. Features

Given the training data (cf. Section II.D) and the biographical information in CBDB, we can create a feature set for each Chinese character in DFZ for training and testing a CRF model. We consider four types of features: original characters, relative positions of named entities in CBDB, whether the character was used in person or location names in DFZ, and whether the characters belong to a named entity.

We explain our features listed in Table 3, using T3, in Section III.B, as a running example.

The original Chinese characters are basic features, summarized in groups 1 and 2 in Table 3. For the position of "州" (zhou1), "州" is an obvious feature in itself. The surrounding $k$ characters can be included in the feature set as well. If we set $k$ to three, the three characters before and after "州", i.e., "仲" (zhong4), "庸" (yong1), "雷" (lei2), "人" (ren2), "廣" (guang3), and "西"(si1), are included in the feature set.

Relative positions of the closest named entities (**NEs**) are summarized in group 3 in Table 3. We consider four types of NEs: office names, entry methods, reign periods, and time markers, and will record NEs on both sides of the current position. The first three types are just like the office, entry, and nianhao labels that we discussed in Section III.A. The time markers refer to a special way of counting years in China, i.e., Chinese sexagenary cycle (干支, gan1 zhi1), and names of months when they were used. We consider NEs that are within 30 characters on either side of the current position, so a position can have up to eight features of group 3.

In T3, there are three characters between "州" and "中書省都事", so **officeRight@3** would be used as a feature for "州". The label name consists of three parts: the type of NEs, the direction respective to the current position (i.e., Right or Left), and the number of characters between the current position and the NE.

Group 4 is about the usage of the current position. It would be helpful to know the probability of the current character being used in a person name or in a location name. Equation (1) shows the basic formula.

$$\Pr(x \text{ in person names}) = \frac{freq(x \text{ in person names})}{freq(x \text{ in } DFZ)} \qquad (1)$$

In T3, "雷" is used as a character in a location name. We calculated the frequency of "雷" being used in location names, and divided this frequency by the total frequency of "雷" in DFZ. We discretized the probability measure into five equal ranges: [0, 0.2), [0.2, 0.4), [0.4, 0.6), [0.6, 0.8), and [0.8, 1.0]. If the probability of "雷" was used in a location name was 0.45, we would add probLoc@3 for "雷", where 3 means the third interval in the discretized ranges.

**Table 4. Performances of selected CRF models**

|    | Group 1 | | | Groups 1+2 | | | Groups 1+2+4+5 | | | Groups 1+2+3+4+5 | | | Groups 1+2+4+5+6 | | |
|----|-------|--------|-------|-------|--------|-------|-------|--------|-------|-------|--------|-------|-------|--------|-------|
|    | Prec. | Recall | $F_1$ | Prec. | Recall | $F_1$ | Prec. | Recall | $F_1$ | Prec. | Recall | $F_1$ | Prec. | Recall | $F_1$ |
| O  | 0.96  | 0.90   | 0.93  | 0.97  | 0.94   | 0.95  | 0.97  | 0.96   | 0.96  | 0.97  | 0.96   | 0.96  | 0.97  | 0.97   | 0.97  |
| NB | 0.76  | 0.91   | 0.83  | 0.85  | 0.94   | 0.89  | 0.91  | 0.94   | 0.93  | 0.90  | 0.94   | 0.92  | 0.93  | 0.95   | 0.94  |
| NI | 0.78  | 0.85   | 0.82  | 0.86  | 0.91   | 0.88  | 0.91  | 0.92   | 0.91  | 0.89  | 0.91   | 0.90  | 0.93  | 0.93   | 0.93  |
| NE | 0.72  | 0.87   | 0.79  | 0.82  | 0.92   | 0.87  | 0.89  | 0.92   | 0.90  | 0.87  | 0.91   | 0.89  | 0.91  | 0.93   | 0.92  |
| AB | 0.78  | 0.83   | 0.80  | 0.85  | 0.86   | 0.86  | 0.89  | 0.88   | 0.88  | 0.88  | 0.87   | 0.87  | 0.91  | 0.89   | 0.90  |
| AI | 0.48  | 0.73   | 0.57  | 0.71  | 0.84   | 0.77  | 0.80  | 0.86   | 0.83  | 0.75  | 0.86   | 0.80  | 0.83  | 0.89   | 0.86  |
| AE | 0.79  | 0.83   | 0.81  | 0.85  | 0.86   | 0.86  | 0.89  | 0.88   | 0.88  | 0.87  | 0.87   | 0.87  | 0.91  | 0.89   | 0.90  |

Group 5 is also about the usage of the current position. There is a list of well-known Chinese family names, which is commonly called Hundred Family Names (百家姓, bai3 jia1 xing4)[3]. We add a feature to the current position if it is in the list. In T3, "陳" (chen2) is such a character. If a family name has two characters, the features will indicate the positions of the characters, e.g., "歐" (ou1) and "陽" (yang2) in "歐陽" will, respectively, have surname@1 and surname@2 as their features.

Features in group 6 are for four types of the named entities, i.e., **office**, **entry**, **nianhao** (for reign period), and **time** (as we discussed for the features in group 3). In general, historians have more complete information about these key types of NEs in Chinese history, so using specific tags for these NEs may offer stronger signals for nearby person and location names.

When we used group 6 along with other groups, we would not annotate a position with features in groups 1 through 5, if the position is part of certain named entities of group 6. Instead, we would use only the values for group 6. For example, at the beginning of the text in Figure 2, we have "洪武元年楊璟" (hong2 wu3 yuan2 nian2 yang2 jing3), where "洪武" represented a reign period, so both characters would be annotated only by **nianhao**. "楊璟" did not belong to any types of NEs in group 6, so would be annotated with other features.

Features of groups 3 and 6 are related in nature. We will see that using group 6 in places of group 3 led to better performance in our evaluations in the next subsection.

## B. Evaluation: Labeled Data

We evaluated the effectiveness of using linear-chain CRF models for recognizing person and location names in DFZ with the labeled data that was discussed in Section II.D. Given the original labels, we could create feature sets for all characters, and then ran 5-fold cross validations.

We classified the characters into seven categories: NB, NI, NE, AB, AI, AE, and O. We use N and A to denote name and location, respectively. B, I, and E denote beginning, internal, and ending, respectively. O means others. Hence, for example, NB is for the first character of a person name and AE is the last character of a location name.

We ran several experiments for CRF models that considered different combinations of the features that we discussed in Section IV.A. The classification results were measured by standard metrics, i.e., precision, recall, and $F_1$ measure that are very common for information retrieval.

Table 4 shows the experimental results for five such combinations. The first row of Table 4 lists the combinations of features used in the experiments. The second row shows the abbreviated names of the performance measures. The leftmost column shows the seven categories of the classification results. The results improved gradually for the experiments listed from the left to the right side, except from "Group 1+2+4+5" to "Group 1+2+3+4+5".

The experiments that used only group 1 as the feature had results that were better than we had anticipated. Identifying categories of individual characters in the dataset of Section II.D did not seem to be a very challenging task. We added the second group of features by setting *k* to five. Then, we added group 4, group 5, group 3, and group 6, one by one for the listed experiments.

Features of groups 3 and 6 are qualitatively related, but offer different performances. Experimental results of "Group 1+2+4+5", "Group 1+2+3+4+5" to "Group 1+2+4+5+6" indicated that using features of group 6 in place of features of group 3 lead to better results.

We also set *k* to three and seven, but we did not observe significant differences in the results. Setting *k* to seven provided a bit better result, but the improvement was not statistically significant.

---

[3] http://baike.baidu.com/subview/6559/15189786.htm

Recognizing the seven categories of individual characters was a first order task for our system. Our goal was to identify person names and location names. Hence, we really care about whether a sequence of NB, NI, and NE, for instance, indeed represented a person name.

We conducted such an integrated verification with the best performing model in Table 4, i.e., using groups 1, 2, 4, 5, and 6 as features. A name, either for a person or for a location, must exactly match the original labels, to be considered as a correct classification. For person names, the precision and recall rates are 92.0% and 93.9%, respectively. For location names, the precision and recall rates are 91.0% and 89.5%, respectively. Finding location names is harder than finding person names.

Table 5. Correct proportions of candidate names decrease with the decreasing CRF scores

| zone | correct | expt'd | zone | correct | expt'd |
|---|---|---|---|---|---|
| 1 | 97 | 1746 | 6 | 70 | 1260 |
| 2 | 88 | 1584 | 7 | 77 | 1386 |
| 3 | 90 | 1620 | 8 | 69 | 1242 |
| 4 | 81 | 1458 | 9 | 59 | 1062 |
| 5 | 79 | 1422 | 10 | 59 | 1011 |

*C. Evaluation: Unlabeled Data*

We trained a CRF model, employing feature groups 1, 2, 4, 5, and 6, with all of the labeled data (Section II.D), and evaluated the model with the task of identifying person and location names in the unlabeled data (Section II.A). It is harder to find person and location names in the original texts of the unlabeled data because the textual natures of original texts of Sections II.A and II.D are different.

The evaluation was conducted in two steps. First, we checked the correctness of the person names. Then, we examined the quality of pairs of person and location names.

For the first task, our CRF model identified 17,914 candidate names. We ranked and split these candidates into 10 zones according to their CRF scores, where each zone contained 1800 candidate names except the last zone which contained the remaining 1714 names.

We manually checked the correctness of the first 100 names in each zone, and recorded the proportion of correct names. A candidate name that was perfectly correct got one point, and a partially correct name got no point. Namely, for a person name with three characters, matching two characters would not receive partial credits.

The "correct" columns in Table 5 list the number of verified names among the 100 samples in every zone. The statistics suggest that the correct proportions decrease with decreasing CRF scores, which is normal for most classification experiments. The range in which we had 80% or better correct rates can be useful for practical historical studies. In this experiment, we achieved 81% for zone 4 and 79% for zone 5. It should be economically worthwhile to manually check all candidate names for real names in the zones with higher correct proportions.

Using the correct rates to estimate the expected number of correct names in the zones[4], we obtain the statistics in columns with headings "expt'd". The sum of the "expt'd" columns is 13,791, meaning that the overall proportion of correct names may be around 77.0%[5]. This estimation method is similar to the area-under-the-curve (**AUC**) method that is used in many machine-learning research papers. The correct rate of 77.0% is much smaller than what we achieved for the labeled data (cf. last paragraph in Section IV.B).

For the second task, we combined a person name and a nearby location name into a pair. The location name must appear after the person name with no more than 10 characters in between.

We created 9148 pairs with this procedure, and found that the locations in 1363 pairs are the native places of the persons in CBDB. These 1363 pairs are known to CBDB, so we did not find new information, but this amount supported the reliability of our methods. For the remaining 7785[6] pairs, the person names (but not the location names) in 2737 pairs were recorded in CBDB, and names in 5048 pairs were new to CBDB. For the former cases, knowledge about history and historical geography are necessary to judge whether the pairs provided new information about a known person name because location names may change over time.

For the latter cases, we can check the DFZ descriptions to verify whether we found new information. Similar to what we did for the first task in this subsection, we ranked and split these 5048 excerpts according to the CRF scores of the names in the original pairs. We split the excerpts into 10 zones, with each zone having 500 excerpts except that the last contained 548 excerpts. We then manually checked (1) whether the names were correct, (2) whether the names and the locations were related, and (3) whether the locations are native places for the names of the first 100 instances in each zone. In Figure 5, we show the correct rates for these three checking facets, respectively, with the **name**, the **name-addr**, and the **name-birth** curves.

---

[4] 1800×0.97=1746; 1800×0.88=1584; etc.
[5] 13791÷17914 = 0.7698
[6] 9148-1363=7785

The curves in Figure 5 still suggest that the correct rates decreased with decreasing CRF scores. Using the AUC estimation method, the overall correct rates for name, name-addr, and name-birth are, respectively, 83.3%, 80.2%, and 52.0%.

The information about a person and his/her native place is very useful for historical studies. Although 52.0% may seem to be a small percentage (cf. the name-birth curve) for most machine-learning researchers, 52.0% of 5048 records represent an attractive amount for historical studies. It would take a lot of human resources to find out 2625 candidate records of person names and their birth places from 900 thousand characters of literary Chinese (cf. Figure 1).

Furthermore, our goals include mining the social networks for historical studies. Finding the relationships between person names and location names can contribute to this goal. For this aspect, more than 80% of the records in the first six zones (cf. the name-addr curve), which have relatively higher CRF scores, are potentially useful as well.

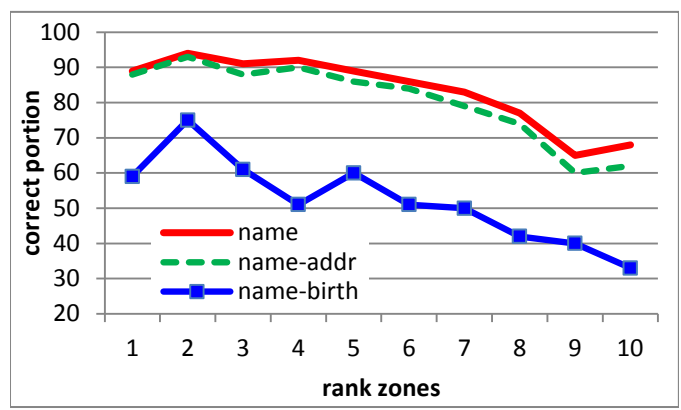

Figure 5. Relatedness between names and addresses

## V. IDENTIFYING PARAGRAPH BOUNDARIES

We evaluated the method for identifying paragraph boundaries that we postulated at the beginning of Section III.D, and checked whether the 3765 records that we presented in Table 2 could lead us to find the boundaries.

To check whether we can obtain a paragraph by the name records, we need a pair of contiguous records. If both records (e.g., "也兒吉尼字尚文" and "陳瑜自仲庸" in Figure 1) are the beginning of two neighbor paragraphs, then we can say that the pair helps us identify a paragraph.

We randomly sampled 205 records from the 3765 records reported in Table 2[7], and referred to them as "**first names**". We also identified the records that immediately followed the first names, and referred to them as the "**second names**". For simplicity of presentation, we use "**name pair**" to refer to the pair of the first and the second names. We wish that a name pair could function like a clipper to identify a paragraph.

In the evaluation, we recorded whether the first name and the second name indeed marked the beginnings of paragraphs. We will denote the percentage that the first and the second names marked the beginnings of a paragraph as **X1** and **X2**, respectively. It is possible that the first and second names of a name pair did not mark paragraph beginnings at the same time, so we recorded the proportion, **X3**, when they simultaneously marked the beginnings.

We also recorded the percentages of when the first and second names lead to the finding of a paragraph. Even when they marked the beginnings of paragraphs, they might not serve our purpose of finding a paragraph when there were two or more paragraphs between them. Recall that we are not really sure that the first and the second names are actual names. We kept two percentages, conditional on whether or not the first and the second names marked the beginnings of paragraphs. The first percentage, denoted by **Y1**, measures the portion of finding a correct paragraph with all name pairs. The second, denoted by **Y2**, is for when only the first and the second name actually marked the beginnings of paragraphs at the same.

Among the 205 name pairs that we sampled, only 199 are good samples. Although we have the text files, we do not have the original images of the text files, while reading the page images is the only way to verify our answers at this moment. Consequently, we could not verify whether we found paragraphs with six samples.

Using the remaining 199 name pairs, we achieved 93.5%, 92.5%, 85.4%, 51.8%, and 58.2% for X1, X2, X3, Y1, and Y2, respectively. The values of X1 and X2 indicated that patterns P5 and P6 are very good markers for paragraph beginnings. The precision rates are high. The value of X3 shows that it was relatively harder for a name pair to simultaneously mark the beginnings of paragraphs. The values of Y1 and Y2 revealed that using P5 and P6 can help us find paragraphs, but there are chances for us to improve the current results.

Many times, we found that there were multiple paragraphs residing between the paragraphs marked by the name pairs. These interrupting paragraphs began with other patterns of biographical information, e.g., "葉溥浙江龍泉人", where "葉溥" is a person name, "浙江" and "龍泉" are addresses, and "人" serves as a key post-marker for native places. An interrupting paragraph may also begin by "楊嘉至正間教諭", where "楊嘉" is person name, "至正" is a reign period, and "教諭" is an office name.

---

[7] This is about 5% sampling rate: 205÷3765≈5.44%.

We could rely on historical knowledge that we extracted from CBDB to annotate the test data with labels for office name, reign period, etc. Then, along with the person names and style names that our programs identified, we could use more patterns, like P8, P9, and P10 listed below, to mark the beginnings of paragraphs.

    **P8: name-address-"人"**
    **P9: name-address-address-"人"** **(e.g.,** "葉溥" -"浙江"- "龍泉"- "人")
    **P10: name-nianhao-office (e.g.,** "楊嘉至正間教諭") **(e.g.,** "楊嘉" -"至正"- "教諭")

After expanding the marking patterns, we found 7972 candidates of paragraph beginnings. Repeating the previous procedure for evaluation, we had 371 effective name pairs and checked whether they helped us identify paragraphs.

We achieved 88.7%, 83.8%, 70.7%, 56.3%, and 73.0% for X1, X2, X3, Y1, and Y2, respectively. The values for X1 and X2 were lowered in this evaluation. This should not be very surprising because we introduced flexible procedures to identify more paragraph beginnings. The current methods spotted more than double the candidate paragraph beginnings than the previous methods (from 3675 to 7972) at the cost of as much as 9% reduction of X2 (from 92.5% to 83.8%). For an analogous reason we observed a reduced X3.

The drops in X1 and X2 were rewarded by a slightly improved Y1 (from 51.8% to 56.3%). Identifying paragraphs with unverified name pairs remained as an imperfect but a possible choice. In the current experiment, we might find 4488 (i.e., 7972×0.563) paragraphs, which is not easy to achieve with only human inspection.

The value of Y2 was significantly improved from 51.8% to 73.0% in this evaluation. After adding the marking patterns, many of the interrupting paragraphs that we failed to spot in the previous evaluation were identified this time.

The value of the multiplication of X3 and Y2 is a good indicator of how often we identified individual paragraphs after spotting the paragraph beginnings. This product was improved from 49.7% to 51.6%.

## VI. DISCUSSIONS AND CONCLUDING REMARKS

We reported our work for mining biographical information from *Difangzhi* with techniques of language models and conditional random fields. Results observed in practical evaluations proved the effectiveness of our methods for named entity recognition and paragraph boundary identification in DFZ.

As illustrated in Figures 1 and 2, processing texts of literary Chinese with computer programs is challenging. We approach this problem with gradually more complex methods. Building our current work on the data that were labeled in previous work [1][19] and CBDB, we were able to apply LM and CRF based models. CNGRAM (Section III.A) is an interactive procedure that was designed to guide researchers to find useful textual patterns.

For practical applications, the LM and CRF models may be integrated with an online tagging service, e.g., MARKUS [8][8], from Leiden University. As we collect more information about person names, style names, pen names, location names, and birthplaces, we become more competent to separate the continuous Chinese strings into paragraphs (cf. Section V) and find social networks of the government officers [3].

We ran our experiments on only 2.4 million characters of *Difangzhi* data, which is a small.[9] However, algorithmically reading and understanding texts of literary Chinese of this scale is very challenging. In the fields of automatic analysis and information retrieval from texts of literary Chinese, we can claim that we are working on a relatively big text collection. After we gather more experience based on the current dataset, we should be able to gradually expand the scale of our experiments. The National Digital Library of China has a dedicated project for collecting *Difangzhi*[10], claiming that more than 400 thousand page images are in the library. We wish we would be able to acquire the parts, in text format, that are related to government officers.

In the near future, we plan to employ more domain-dependent knowledge and contextual constraints to recognize and disambiguate named entities. People of different dynasties may have the same name, for instance. In *Difangzhi*, records about government officers of the same dynasty usually appear close to each other. Many times, the records were ordered chronically. Considering these practical constraints for disambiguation can make our annotations about a person more precise.

In the long run, mining the grammar rules of literary Chinese is a bigger and more rewarding challenge. It was found that the language models and CRF models worked better for some of the 83 *Difangzhi* than others [3]. People who compiled these books adopted different styles of writing, and the styles varied from time to time and from place to place. Knowing the grammar rules that govern these language patterns will enable us to find more precise information from *Difangzhi* and perhaps other historical documents written in literary Chinese.

---

[8] http://chinese-empires.eu/analysis/tools/
[9] This paragraph is added to the paper for responding to a reviewers' comment.
[10] National Digital Library of China: http://mylib.nlc.gov.cn/web/guest/zhengjidifangzhi (accessed on 4 October 2015)

## ACKNOWLEDGEMENTS

This work was supported in part by the Ministry of Science and Technology (MOST) of Taiwan under grants 102-2420-H-004-054-MY2 and 104-2221-E-004-005-MY3. We thank the anonymous reviewers for their valuable comments and pointers.

## REFERENCES

[1] P. K. Bol. Historical research in a digital environment, keynote speech in the 3rd International Conference on Digital Archives and Digital Humanities, 2012. <http://isites.harvard.edu/fs/docs/icb.topic1080143.files/Historical%20Research%20in%20Dig%20Env.ppt>.

[2] P. K. Bol., J. Hsiang, and G. Fong. Prosopographical databases, text-mining, GIS and system interoperability for Chinese history and literature, *Proceedings of the 2012 International Conference on Digital Humanities*, 2012.

[3] P. K. Bol, C.-L. Liu, and H. Wang. Mining and discovering biographical information in *Difangzhi* with a language-model-based approach, Presented in the 2015 International Conference on Digital Humanities, 2015.

[4] P. Boonkwan and M. Steedman. Grammar induction from text using small syntactic prototypes, *Proceedings of the 5th International Joint Conference on Natural Language Processing*, 438–446, 2011.

[5] C. de la Higuera. A bibliographical study of grammatical inference, *Pattern Recognition*, 38:1332–1348, 2005.

[6] J. Gao, M. Li, A. Wu, and C.-N. Huang. Chinese word segmentation and named entity recognition: A pragmatic approach, *Computational Linguistics*, 31(4):531–574, 2005.

[7] J. M. Hargett. Song dynasty local gazetteers and their place in the history of *Difangzhi* writing, *Harvard Journal of Asiatic Studies*, 56(2):405–442, 1996.

[8] H. I. Ho. MARKUS: A fundamental semi-automatic markup platform for classical Chinese, Presented in the 2015 International Conference on Digital Humanities, 2015.

[9] H.-H. Huang, C.-T. Sun, and H.-H. Chen. Classical Chinese sentence segmentation, *Proceedings of CIPS-SIGHAN Joint Conference on Chinese Language Processing*, 15–22, 2010.

[10] L. Huang, Y. Peng, H. Wang, and Z. Wu. PCFG parsing for restricted classical Chinese texts, *Proceedings of the 1st SIGHAN Workshop on Chinese Language processing*, 1–6, 2001.

[11] R. Hwa. Supervised grammar induction using training data with limited constituent information, *Proceedings of the 37th Annual Meeting of the Association for Computational Linguistics*, 73–79, 1999.

[12] Y.-C. Kuo. *Using Reinforcement Learning to Learn Phrase Structure Parsing in Mandarin Chinese*, Master's Thesis, National Tsing Hua University, Taiwan, 2009. (in Chinese)

[13] X.-X. Lai (來新夏), *China Difangzhi* (中國地方志), Taiwan Commercial Press, 1995. (in Chinese)
<https://books.google.com.tw/books?id=j4zXTfWWVTYC>

[14] J. Lee and Y. H. Kong. A dependency treebank of classical Chinese poems, *Proceedings of 2012 Conference of the North American Chapter of the Association for Computational Linguistics: Human Language Technologies*, 191–199, 2012.

[15] C.-L. Liu, C.-K. Huang, H. Wang, and P. K. Bol. Toward algorithmic discovery of biographical information in local gazetteers of ancient China, *Proceedings of the 29th Pacific Asia Conference on Language, Information and Computation*, 87–95, 2015.

[16] Y. Lü, S. Li, T. Zhao, and M. Yang. Learning Chinese bracketing knowledge based on a bilingual language model, *Proceedings of the 19th International Conference on Computational Linguistics*, 1–7, 2002.

[17] A. K. McCallum. MALLET: A Machine Learning for Language Toolkit, 2002. <http://mallet.cs.umass.edu>

[18] D. Nadeau and S. Sekine. A survey of named entity recognition and classification, *Linguisticae Investigationes*, 30(1):3–26, 2007.

[19] W.-h. Pang, S.-p. Chen, and H. Cheng. From text to data: Extracting posting data from Chinese local monographs, *Proceedings of the 5th International Conference on Digital Archives and Digital Humanities*, 93–116, 2014.

[20] C. Sutton and A. McCallum. An introduction to conditional random fields, *Foundations and Trends in Machine Learning*, 4(4):267–373, 2011.

[21] D. Yarowsky. Unsupervised word sense disambiguation rivaling supervised methods, *Proceedings of the 33rd Annual Meeting of the Association for Computational Linguistics*, 189–196, 1995.

[22] K. Yu, Y. Miyao, X. Wang, T. Matsuzaki, and J. Tsujii. Semi-automatically developing Chinese HPSG grammar from the Penn Chinese Treebank for deep parsing, *Proceedings of the 23rd International Conference on Computational Linguistics*: posters, 1417–142, 2010.